\documentclass[11pt]{article}
\newif\ifarxiv
\arxivtrue

\usepackage{acl}

\usepackage{xurl}
\usepackage{times}
\usepackage{latexsym}
\usepackage[T1]{fontenc}
\usepackage[utf8]{inputenc}
\usepackage{inconsolata}
\usepackage{graphicx}
\graphicspath{{figures/}}
\usepackage{placeins}
\usepackage{booktabs}
\usepackage{tabularx}
\usepackage{ragged2e}
\usepackage{array}
\usepackage{threeparttable}
\usepackage{enumitem}
\usepackage{amsmath}
\usepackage{amssymb}
\usepackage{longtable}
\DeclareUnicodeCharacter{2013}{--}
\DeclareUnicodeCharacter{2014}{--}
\DeclareUnicodeCharacter{2019}{'}
\DeclareUnicodeCharacter{2018}{`}
\DeclareUnicodeCharacter{201C}{``}
\DeclareUnicodeCharacter{201D}{''}
\DeclareUnicodeCharacter{2026}{\ldots}
\DeclareUnicodeCharacter{00A0}{~}
\DeclareUnicodeCharacter{2265}{\ensuremath{\geq}}
\DeclareUnicodeCharacter{2248}{\ensuremath{\approx}}
\DeclareUnicodeCharacter{2212}{\ensuremath{-}}
\DeclareUnicodeCharacter{03BA}{\ensuremath{\kappa}}
\DeclareUnicodeCharacter{0394}{\ensuremath{\Delta}}
\DeclareUnicodeCharacter{03C6}{\ensuremath{\phi}}
\DeclareUnicodeCharacter{00D7}{\ensuremath{\times}}
\DeclareUnicodeCharacter{2713}{\checkmark}
\DeclareUnicodeCharacter{2717}{x}
\title{Conflict or Strategy? Asymmetric Role Framing of La France insoumise and Rassemblement National in French News Headlines, 2022–2025}

\author{Amr Sobhy \\ Le French News Lab \\ \texttt{amr@frenchnewslab.org}}

\begin{document}
\maketitle

\begin{abstract}

Do French news headlines frame left- and right-populist challengers as symmetric "extremes," or as fundamentally different political adversaries? We examine 28,592 headlines about La France insoumise (LFI) and Rassemblement National (RN) published by 25 French-language outlets between 2022 and 2025, annotated through a three-model LLM pipeline validated against a stratified human audit. The clearest finding is role asymmetry rather than valence asymmetry: conflict framing and strategic-game framing are more robust across models and time than delegitimization, with AGGRESSOR serving as corroborating role syntax. LFI appears in headlines more often through a conflict register and RN through a strategic-electoral register. This role gap is direction-stable across all three annotation models, survives bootstrapping and permutation tests, and persists across outlet families and most of 2022–2025. Because the design is observational, we report how headlines distribute these roles, not evidence that the press constructs them independently of what the parties did. A secondary moral-accounting layer (who is blamed, legitimized, or cast as a victim) is structured by outlet rather than party, producing aggregate nulls that conceal some of the corpus's most polarized patterns. Methodologically, the annotation pipeline reveals a two-tier reliability profile: conflict and strategic-game framing achieve the strongest human validation and cross-model stability; actor role is direction-stable but treated as corroborating because its audit reliability is lower; normative-judgment constructs (legitimacy, blame) are weaker. The paper contributes political-role assignment as a target for computational framing research that decomposes what valence-based measures conflate, and establishes a construct-stratified reliability framework for calibrating majority-vote LLM annotation pipelines in political text tasks.

\end{abstract}
\section{Literature Review and Contributions}

\citet{entman1993}'s definition of framing as a dual operation of \emph{selection} and \emph{salience} underlies this paper. Agenda-setting tracks selection through coverage volume \citep{mccombsShaw1972,shoemakerVos2009}; research on framing devices tracks how worded devices such as word choice, metaphors, and exemplars contribute to frame salience \citep{gamsonModigliani1989,gravesSandriman2026}. We add a third layer: \emph{political-role assignment}, meaning what kind of political actor a party is made to appear as across repeated coverage (aggressor, strategic competitor, blamed actor, legitimate or illegitimate participant). Role-adjacent constructs are present in \citet{entman1993}'s original definition (problem definition and causal interpretation each imply an actor's assigned role), but the explicit operationalization of asymmetric political-role annotation at headline scale in French is, to our knowledge, new. This moves analysis beyond generic negativity to asymmetric role construction. Headlines are the analytic unit: short, widely circulated, and known to shape memory and inference \citep{ecker2014,gabielkov2016}.

\emph{For NLP readers}: the political-role annotation used here operates at the discourse level (assigning a role to a named party across a full headline) and differs from sentence-level semantic role labeling (SRL). SRL identifies argument structure around a predicate; our scheme identifies the political standing a party is given across the editorial framing of an event, requiring pragmatic and contextual cues that syntactic SRL parsers are not designed to capture. NLP has studied discourse-level political framing through entity framing and role portrayal \citep{mahmoud2025entity}; our contribution is large-scale French application with cross-model majority-vote validation that separates construct tiers by cross-model stability.

Computational framing has expanded rapidly but remains one-dimensional. Sentiment and bias classifiers (BASIL, \citealt{fan2019}; BABE, \citealt{spinde2021}) detect evaluative language; issue-frame corpora such as the Media Frames Corpus \citep{card2015} and shared-task resources that annotate genre, framing, and persuasion techniques in online news (SemEval-2023 Task 3, \citealt{piskorski2023}) extend that space further. Neither cleanly separates selection from salience, and almost none capture the \emph{political role} assigned to a named party---a dimension the field's own theorists require but which, as \citet{otmakhova2024}'s ACL survey documents, NLP approaches have systematically underspecified. For French specifically, broad issue-frame lexicons remain insufficient for role-level coding, and the French-language resource base for party-level framing at headline scale is, to our knowledge, largely absent; NLP has addressed adjacent problems such as political sentiment and media bias detection in English-language news \citep{fan2019,spinde2021}, but without the party-level role-assignment focus this paper targets.

\emph{Conflict} framing is a well-established generic register of political news coverage \citep{semetko2000}. In our annotation scheme, we treat it as logically separable from \emph{strategic-game} framing: conflict portrays behavioral struggle, whereas strategic-game portrays electoral competition. Both can attach asymmetrically to challengers of different ideological color. Left- and right-populisms differ in the targets of their identity logics \citep{mudde2017}, but to our knowledge no existing computational design places both inside the same media system to test whether the press frames them through symmetrical or divergent roles. The French press is politically plural and historically structured in closer relation to the political field \citep{bensonHallin2007,cage2022}. We ask whether French headlines frame LFI and RN as the same kind of political object, which roles are assigned to each party, and where outlet-specific narratives reverse or intensify the aggregate pattern. Because normative-judgment constructs (e.g., legitimacy) show high cross-model disagreement and sensitivity to outlet ideology, we restrict our primary empirical claims to conflict and strategic-game framing, using AGGRESSOR as corroborating role syntax after party-stratified validation.

\section{Research Questions}

Five questions organize the analysis. First, do French headlines frame LFI and RN as different kinds of political adversaries rather than as symmetric "extremes"? Second, which behavioral roles are assigned to each party through conflict framing, strategic-game framing, and aggressor positioning? Third, how do legitimacy, blame, normalization, and target status modify or contest those behavioral roles? Fourth, are these asymmetries stable across years and annotation models? Finally, where do outlet-specific narratives reverse or intensify the aggregate pattern?

\section{Data and Methods}

\subsection{Corpus}

The corpus consists of \textbf{28,592 French news headlines} drawn from a source database of 902,111 headlines published by 25 French-language outlets between January 2022 and December 2025.Headlines were included if they contained at least one surface reference to La France Insoumise (LFI), a radical-left populist party, the Rassemblement National (RN), a nationalist radical-right party, LFI-led electoral coalitions, or selected high-salience party leaders: Jean-Luc Mélenchon, Mathilde Panot, and Manon Aubry for LFI; Marine Le Pen and Jordan Bardella for RN.\footnote{Full search terms, coalition-retention rules, and exclusion decisions appear in Appendix~\ref{app:corpus}.} Because broad coalition terms (e.g., NUPES) risk injecting non-LFI coverage, they were subjected to a strict majority-vote primary-target filter; Appendix~\ref{app:corpus} confirms this retained coalition subset does not artificially drive the observed LFI conflict tally. Of the 28,592 headlines, 12,741 primarily frame LFI, 14,770 primarily frame RN, and 1,081 frame both parties jointly or set them in explicit opposition (coded \emph{BOTH}). The corpus spans four calendar years, with the heaviest coverage in the 2024 election year (\emph{n} = 9,697) and the lightest in the non-election year 2023 (\emph{n} = 4,652).

BOTH headlines are excluded from the main LFI-versus-RN regressions so each observation has a single primary party target. A three-way stability check confirms the exclusion does not inflate or suppress the core role asymmetry: all eight primary outcomes shift by at most $|\Delta\mathrm{OR}| \leq 0.004$ with no direction reversals when BOTH headlines are included in a three-level sensitivity model (full specification and robustness checks in Appendix~\ref{app:both-sensitivity}). The BOTH sub-corpus is otherwise retained in corpus description and validation; its elevated conflict rate (24.2\%, nearly double LFI's 13.8\%) and strategic-game rate (59.3\%, highest in the corpus) are consistent with comparative electoral coverage driving joint mentions.

Appendix Table~\ref{tab:appendix-corpus} describes the corpus by outlet, including approximate editorial orientation and per-party headline counts. Outlets range from centrist and left-leaning press (\emph{Le Monde}, \emph{Libération}, \emph{L'Humanité}) to right-leaning and far-right publications (\emph{Le Figaro}, \emph{Valeurs actuelles}, \emph{Causeur}, \emph{Fdesouche}), with large-circulation television and digital outlets (\emph{Franceinfo}, \emph{BFMTV}, \emph{TF1 INFO}, \emph{Le Parisien}, \emph{20 Minutes}) spanning the middle.\footnote{\emph{Fdesouche} is a far-right aggregator republishing others' headlines ($<$2.5\% of the corpus; 690 rows), absorbed into the residual ``other'' category in all cluster-bootstrap models. Excluding its 645 LFI/RN rows leaves all four main outcome directions unchanged ($\mathrm{OR}$ shifts $\leq 0.034$).} Orientation labels are descriptive author classifications grounded in established French media-system typologies and audience-trust work \citep{bensonHallin2007,cevipof2019barometre}, not modeled as ground-truth ideology; the inferential unit is the outlet-specific interaction, not the orientation label itself. This diversity matters: no single outlet's stance can explain aggregate LFI-RN differences, and outlet-level heterogeneity is directly estimable.

\subsection{Annotation Scheme}

The annotation scheme has a strict hierarchy: conflict and strategic-game framing carry the strongest behavioral claims; actor role supplies corroborating syntax; legitimacy supplies a supporting moral-accounting layer; blame, target, reference form, personalization, and party target are diagnostic fields. The boundary rules are framed around the distinct question each construct asks of the same headline. A conflict frame identifies behavioral confrontation; actor role identifies syntactic position; strategic-game framing identifies the electoral register; legitimacy identifies institutional standing; and blame identifies causal or moral responsibility for attributable harm.

The foreground constructs are the binary conflict frame and binary strategic-game frame. The four-way actor-role field (AGGRESSOR / TARGET / NEUTRAL / MIXED) is retained as corroborating syntax: its direction is stable and party-stratified audit errors are symmetric, but its overall validation is weaker than the two binary behavioral fields. Legitimacy frame (NORMALIZING / DELEGITIMIZING / NONE) distinguishes behavioral role assignment from moral accounting; lower model agreement (53.4\% unanimous) and contested human-audit performance mean it is treated as a \emph{structured tendency} rather than a stable property. Blame, reference form, and personalization are diagnostic fields: blame because its substantive payoff appears in outlet interactions, and reference form/personalization because they indicate whether coverage is routed through parties or leaders. Party target is used for corpus filtering and party-level analysis, not as an outcome variable.

 \begin{table*}[t]
\centering
\scriptsize
\setlength{\tabcolsep}{4pt}
\renewcommand{\arraystretch}{1.08}
\begin{tabularx}{\textwidth}{@{}l l >{\RaggedRight\arraybackslash}X >{\RaggedRight\arraybackslash}X@{}}
\toprule
Field & Type & Categories / values & Construct \\
\midrule
\texttt{conflict\_frame} & Binary & TRUE / FALSE & Active adversarial framing: attack, clash, denunciation \\
\texttt{blame\_frame} & Binary & TRUE / FALSE & Explicit causal or moral responsibility for harm or failure \\
\texttt{strategic\_game\_frame} & Binary & TRUE / FALSE & Electoral competition, tactical positioning, alliance arithmetic \\
\texttt{legitimacy\_frame} & 3-way & NORMALIZING / DELEGITIMIZING / NONE & Normative positioning of the party \\
\texttt{actor\_role} & 4-way & AGGRESSOR / TARGET / NEUTRAL / MIXED & Active or passive adversarial role \\
\texttt{reference\_form} & 3-way & PARTY / LEADER / MIXED & Unit of political reference (organization vs. individual) \\
\texttt{party\_target} & 4-way & LFI / RN / BOTH / UNCLEAR & Primary framing target (diagnostic) \\
\texttt{personalization} & 3-way & PARTY\_CENTRIC / LEADER\_CENTRIC / MIXED & Dominant unit of coverage framing (diagnostic) \\
\bottomrule
\end{tabularx}
\caption{Annotation scheme. Full codebook definitions and tie-break rules in Appendix~\ref{app:codebook}; guidelines follow established content-analysis practice \citep{neuendorf2017}.}
\label{tab:annotation-scheme}
\end{table*}

\subsection{Annotation Pipeline: Three-Model Majority Vote}

Each headline was independently annotated by three open-weight large language models served through Regolo (\url{https://regolo.ai}), a cloud inference API: GPT-OSS-120B \citep{openai2025gptoss}, Llama-3.3-70B-Instruct \citep{meta2024llama33}, and Mistral Large 2 (123B) \citep{mistral2024large2}, all queried with an identical annotation prompt specifying the codebook definitions and tie-break rules.\footnote{Model identifiers as recorded: \texttt{gpt-oss-120b}, \texttt{Llama-3.3-70B-Instruct}, Mistral Large 2 (123B). API access dates are reported in the supplementary materials.} For binary fields, the final label is the majority vote of the three models. For categorical fields with more than two values, the final label is the plurality label (two or more models agreeing); in the small proportion of cases where all three models assign different categorical values (\emph{three-way splits}), the label is flagged and treated as ambiguous in all downstream analyses.

Three prompt-design choices are central to the pipeline's portability as a Computational Social Science (CSS) contribution. First, behavioral and normative constructs are operationalized in natural language through role-specific boundary conditions: \texttt{conflict\_frame} is TRUE only when the headline contains an explicit confrontation verb directed at a named actor (e.g., \emph{attaque, dénonce, cible} [attacks, denounces, targets]) and FALSE for emotional reactions, attributed quotes, or structural political reporting without a direct confrontation verb. This distinction systematically excludes sentiment from the behavioral register. Second, the \texttt{legitimacy\_frame} tie-break rule for three-way categorical splits instructs models to return NONE when all three values disagree (rather than taking a forced plurality from a single model), so three-way split rows are treated as ambiguous throughout, making the calibration floor explicit rather than obscuring it in the aggregate. Third, negative examples are embedded in the prompt for the two highest-disagreement fields (\texttt{conflict\_frame} and \texttt{legitimacy\_frame}) to anchor models to construct boundaries rather than surface valence cues; full examples appear in Appendix~\ref{app:prompt-excerpt}. These choices maximize the pipeline's replicability in other party systems, not only its accuracy on LFI/RN.

Model calls were made independently, without exposing any model to the others' outputs. The three models represent different providers and architectural families (GPT-OSS, Meta's Llama series, and Mistral AI's large-parameter series), which is a necessary but not sufficient condition for independence.

Table~\ref{tab:reliability} reports inter-model and human inter-coder agreement for all main framing fields. The binary behavioral constructs show the highest model unanimity (79.6–85.5\%). \texttt{legitimacy\_frame} is the most contested field at 53.4\% unanimous (42.8\% of its corpus-level labels are 2-of-3 majority labels), which is why delegitimization claims rest on a narrower evidentiary base than the behavioral-role findings.

We treat LLM annotation as a \textbf{robust labeling procedure} \citep{heseltine2024,tornberg2025}, not as ground truth: majority vote eliminates dependence on any single model's idiosyncrasies but does not eliminate shared bias from training-data overlap. Frontier LLMs can match expert coding on political text when prompts are explicit \citep{heseltine2024,tornberg2025}, though political annotation is vulnerable to party-cue effects \citep{vallejo2025}. Three safeguards address correlated bias: per-field agreement rates (Table~\ref{tab:reliability}), explicit uncertainty treatment for contested fields, and the 400-headline human audit (§3.4). Findings are graded accordingly: foreground claims require stability across these safeguards; structured-tendency claims survive with caveats.

Per-model regressions (Appendix Table~\ref{tab:per-model-robustness}) confirm behavioral-tier direction-stability across all three models; only Mistral reverses on moral-accounting fields, independently motivating the foreground/structured-tendency distinction. Cluster-bootstrapped CIs and a 10,000-shuffle permutation test corroborate both core asymmetries: zero of 10,000 permutations produced a conflict OR as low as the observed 0.594 (raw covariate-free; fixed-effect adjusted OR = 0.614, Table~\ref{tab:main-results}) or a strategic-game OR as high as the observed 1.431 (raw covariate-free; fixed-effect adjusted OR = 1.414, Table~\ref{tab:main-results}) (both $p < 0.0001$; Appendices~\ref{app:bootstrap}--\ref{app:multiple-corrections}).

\subsection{Human Validation Study}

A stratified 400-headline audit stress-tests the pipeline at its weakest points: disagreement-heavy cases (\emph{n} = 150), high-agreement controls (\emph{n} = 150), and theory-targeted oversamples (\emph{n} = 100). The 1.4\% overall validation rate (400 of 28,592) is justified by the stratified design: the 150-headline disagreement stratum is entirely oversampled relative to its corpus frequency, making this audit more diagnostic than a proportional random draw of the same size. Random seed 42 was frozen before coding. Two annotators (both graduate-level researchers with backgrounds in political communication and French media) annotated all 400 headlines blind to model output after a calibration session on 30 held-out practice headlines (codebook in Appendix~\ref{app:codebook}).
\begin{figure*}[!t]
\centering
\includegraphics[width=\textwidth]{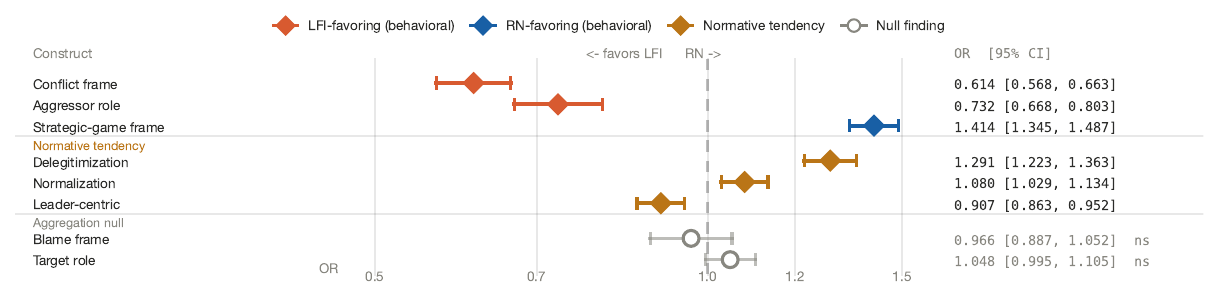}
\caption{Adjusted RN-vs-LFI odds ratios for all eight constructs (party-targeted headlines, $n = 27{,}511$; outlet and year fixed effects), grouped by evidentiary tier: behavioral, normative tendency, and aggregation null. Whiskers: 95\% CI; reference line: OR = 1.0.}
\label{fig:forest}
\end{figure*}

\begin{table}[t]
\centering
\scriptsize
\setlength{\tabcolsep}{5pt}
\renewcommand{\arraystretch}{1.08}
\resizebox{\columnwidth}{!}{%
\begin{tabular}{@{}lrrr@{}}
\toprule
Field & 3/3 agree & Human \% agree & Human $\kappa$ \\
\midrule
\texttt{conflict\_frame}        & 85.5\% & 90.0\% & 0.762 \\
\texttt{strategic\_game\_frame} & 79.6\% & 91.0\% & 0.805 \\
\texttt{actor\_role}            & 68.7\% & 74.0\% & 0.614 \\
\texttt{legitimacy\_frame}      & 53.4\% & 77.5\% & 0.644 \\
\texttt{blame\_frame}           & n/a    & 80.2\% & 0.523 \\
\bottomrule
\end{tabular}
}
\caption{Three-model unanimous agreement (N = 28,592) and human inter-coder $\kappa$ (N = 400; Landis--Koch thresholds, \citealt{landisKoch1977}). \texttt{blame\_frame} excluded from model-agreement pass.}
\label{tab:reliability}
\end{table}

Four of five fields reach $\kappa \geq 0.60$; \texttt{blame\_frame} ($\kappa = 0.523$, moderate) is treated as a supporting indicator throughout. Human–MV agreement is above threshold for all audited fields (Appendix~\ref{app:human-validation}); final three-way \texttt{legitimacy\_frame} splits remain unreliable (human–MV = 31.2\% on 16 split rows), so delegitimization is retained as a \emph{structured tendency}. Stratum-level breakdowns and boundary-case analysis are in Appendices~\ref{app:human-validation}--\ref{appendix-error-analysis}.

\subsection{Statistical Analysis}

Framing rates are compared as proportions (chi-squared tests) and through logistic regression with outlet and year fixed effects, estimating party-level gaps net of compositional differences. For each binary outcome $Y$, the pooled model is $\mathrm{logit}\{P(Y=1)\} = \alpha + \beta\,\mathrm{RN} + \gamma_{\mathrm{outlet}} + \delta_{\mathrm{year}}$; reported odds ratios are $\exp(\beta)$ from standard statsmodels maximum-likelihood logit fits. The interaction models replace the pooled party term with party × outlet or party × year terms and use likelihood-ratio tests against the pooled fixed-effect model. The pooled party comparisons are the pre-specified core tests; outlet × party and year × party models are descriptive follow-ups, and a pattern is described as "outlet-structured" only when at least three outlets show convergent evidence. For interaction models, the analysis estimates eight named outlet families plus an `other` residual category; low-count outlets such as Causeur, Slate.fr, Blast, and Ouest-France are absorbed into `other` rather than reported as standalone outlet coefficients. Effect sizes (odds ratios, Cramér's $V$) are interpreted alongside $p$-values. Three robustness checks (per-model regressions, cluster-bootstrapped CIs [9 outlet families; sub-optimal cluster count; see §\ref{sec:delegit}], permutation test) and a post-hoc lexical analysis supporting §4.3 appear in Appendices D, F--H.

\section{Results}

Results follow the §3 evidentiary hierarchy: foreground claims require model agreement and human validation; structured-tendency claims survive with caveats; outlet-structured claims are weak in aggregate but strong in interaction models.

Table~\ref{tab:main-results} presents the full distribution; it does not support a simple aggregate-negativity claim, and the role-based pattern is developed below.

\begin{table*}[t]
\centering
\scriptsize
\setlength{\tabcolsep}{2pt}
\renewcommand{\arraystretch}{1.08}
\begin{tabularx}{\textwidth}{@{}>{\RaggedRight\arraybackslash}Xrrrr>{\RaggedRight\arraybackslash}Xl@{}}
\toprule
Construct & LFI rate & RN rate & Δ (RN − LFI) & RN OR & 95\% CI & \emph{p} \\
\midrule
\texttt{conflict\_frame} & 0.138 & 0.087 & −0.051 & 0.614 & [0.568, 0.663] & *** \\
\texttt{blame\_frame} & 0.087 & 0.087 & −0.000 & 0.966 & [0.887, 1.052] & ns \\
\texttt{strategic\_game\_frame} & 0.372 & 0.459 & +0.087 & 1.414 & [1.345, 1.487] & *** \\
\texttt{legitimacy\_frame} = DELEGITIMIZING & 0.257 & 0.312 & +0.055 & 1.291 & [1.223, 1.363] & *** \\
\texttt{legitimacy\_frame} = NORMALIZING & 0.437 & 0.451 & +0.014 & 1.080 & [1.029, 1.134] & ** \\
\texttt{actor\_role} = AGGRESSOR & 0.086 & 0.063 & −0.023 & 0.732 & [0.668, 0.803] & *** \\
\texttt{actor\_role} = TARGET & 0.301 & 0.315 & +0.014 & 1.049 & [0.995, 1.105] & ns \\
\texttt{personalization} = LEADER\_CENTRIC / MIXED & 0.514 & 0.477 & −0.037 & 0.907 & [0.863, 0.953] & *** \\
\bottomrule
\end{tabularx}
\caption{Frame rates and logistic regression by party (party-targeted headlines, n = 27,511; BOTH excluded). OR: RN vs.\ LFI, outlet and year fixed effects. Stars are standard MLE/Wald (*** \emph{p} < 0.001; ** \emph{p} < 0.01; * \emph{p} < 0.05; ns = not significant); Wald intervals treat within-outlet observations as independent and are anti-conservative. The stricter outlet-cluster bootstrap (Appendix~\ref{app:bootstrap}) is the inferential anchor: conflict and strategic-game retain significance as the two primary behavioral claims; AGGRESSOR also retains it but is treated as corroborating role syntax (weaker validation); the moral-accounting rows do not retain it. All six significant findings survive BH correction ($q = 0.05$); both nulls remain non-significant (Appendix~I).}
\label{tab:main-results}
\end{table*}

\subsection{The Core Asymmetry: LFI Conflict, RN Strategy}

The strongest stable finding is not generalized negativity but a split between two registers. LFI is covered more through a behavioral conflict register, appearing more often than RN in a conflict frame (13.8\% vs. 8.7\%; +5.1 pp; adjusted RN OR = 0.614, \emph{p} < 0.001). RN is covered more through a strategic-electoral register (45.9\% vs. 37.2\%; adjusted RN OR = 1.414, \emph{p} < 0.001). These two primary behavioral estimates survive the stricter outlet-cluster bootstrap that relaxes the within-outlet independence assumption: conflict 95\% CI [0.572, 0.652], strategic-game [1.226, 1.635] (Appendix~\ref{app:bootstrap}). AGGRESSOR provides corroborating role syntax: LFI is more likely than RN to be cast as the aggressor in a confrontation (8.6\% vs. 6.3\%; adjusted RN OR = 0.732, bootstrap CI [0.636, 0.814]), but this label is interpreted cautiously because actor-role validation is weaker than the two binary behavioral fields and audit errors are symmetric by party (Appendix~\ref{tab:actor-party}). Crucially, AGGRESSOR measures syntactic and behavioral posture (narrating the party as actively initiating a challenge or critique), rather than serving as an editorial verdict of hostility; 91.4\% of LFI headlines and 93.7\% of RN headlines carry no aggressor label. The significance stars on the moral-accounting rows of Table~\ref{tab:main-results} are standard MLE/Wald estimates only and do not survive this bootstrap (§\ref{sec:delegit}, Appendix~\ref{app:bootstrap}); we therefore do not treat them as foreground inferential claims.

\subsection{Delegitimization: A Structured Tendency, Not a Stable Property}
\label{sec:delegit}

The pooled estimate shows RN headlines more likely to carry a delegitimizing frame (31.2\% vs. 25.7\%; adjusted RN OR = 1.291, \emph{p} < 0.001), but four considerations qualify it. First, legitimacy judgments are more sensitive to model disagreement than the paper's behavioral-role findings, with only 53.4\% unanimous agreement in the final ensemble (§3.3). Second, the human validation audit shows that legitimacy is among the more contested audited fields between human coders (κ = 0.644) and the lowest-agreement field between humans and the final majority vote (74.5\% overall, 31.2\% on final three-way model splits). Third, the cluster-bootstrapped confidence interval for delegitimization (Boot 95\% CI: [0.999, 2.003]; Appendix~\ref{app:bootstrap}) is 7.17$\times$ wider than the Wald interval and nearly includes 1.0. This reflects the nine bootstrap clusters being below the standard $\geq$30 threshold and the high intra-cluster correlation for legitimacy judgments.\footnote{Both tiers share the same sub-optimal cluster count; the behavioral tier retains bootstrap significance due to effect size, not better calibration.} Fourth, the temporal decomposition (§4.3) shows the gap reversing year-on-year. We therefore frame RN delegitimization as a \emph{structured tendency} organized by outlet and electoral period, not as a stable property of French press coverage. This outlet-structured pattern is consistent with theoretical accounts of far-right party normalization, especially work documenting how d\'ediabolisation repositions the party for mainstream electoral legitimacy \citep{ivaldi2014}.

More substantively, delegitimization is among the most clearly outlet-structured dimensions in the dataset. Table~\ref{tab:outlet-interactions} reports outlet-level ORs for all three moral-accounting dimensions jointly. RN delegitimization (raw rates: \emph{Libération} 48.9\% vs.\ 20.4\% LFI; \emph{JDD} 21.6\% vs.\ 36.4\% LFI) is concentrated in left-leaning outlets and reverses at \emph{JDD} and \emph{Le Figaro}; there is no uniform French media stance toward RN legitimacy.

\subsection{Temporal Structure: Strategic Framing Peaks, Conflict Persists}

The main asymmetries do not behave uniformly over time (year-level raw rates in Appendix~\ref{app:lexical}; Figure~\ref{fig:temporal} plots the trajectories).

\begin{figure}[htb]
\centering
\includegraphics[width=0.88\columnwidth]{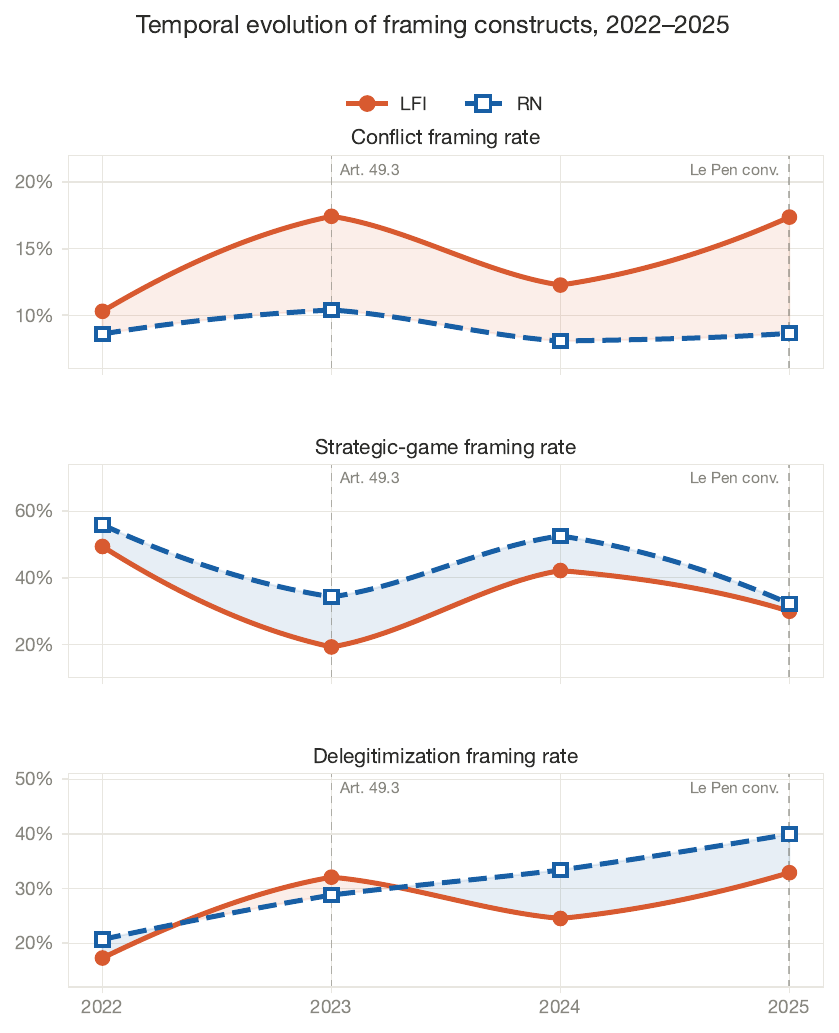}
\caption{Year-level framing rates for LFI (coral) and RN (blue), 2022--2025. Shading: higher-rate party; dashed verticals: Article~49.3 (2023) and Le~Pen's conviction (2025). Conflict (top) persists; strategic-game (middle) peaks in 2023 and collapses in 2025; delegitimization (bottom) reverses in 2023 then returns RN-positive.}
\label{fig:temporal}
\end{figure}

RN's strategic advantage is strongest in 2023 (RN-LFI gap = +0.150; OR = 2.205) and remains visible in 2024, but it essentially disappears in 2025 (gap = +0.022; OR = 1.087, ns). By contrast, LFI's conflict excess is present in every year and is strongest in 2025 (gap = −0.087; OR = 0.464). Delegitimization is the least temporally uniform result: it reverses in 2023 before returning to an RN-positive gap in 2024 and 2025.

A post-hoc lexical-shift diagnostic supports both anomalies (full outputs in Appendix~\ref{app:lexical}): LFI delegitimizing headlines in 2023 overrepresent pension-reform and Hamas/Israel terms, while RN headlines in 2025 shift away from electoral-competition terms toward judicial and ineligibility vocabulary.

The 2023 reversal is best understood as a crisis-specific interruption: the pension-reform conflict moved LFI into a high-volume legitimacy register as it led parliamentary and street opposition to the government's use of Article 49.3. Meanwhile 2023 was a lower-volume non-election year for RN, reducing routine RN delegitimization and making LFI's crisis posture unusually salient.

The 2025 strategic-game compression has a different source: Marine Le Pen's March 2025 conviction shifted RN coverage toward judicial and eligibility registers, displacing ordinary electoral-positioning language. The result does not invalidate the core role asymmetry; it shows that the strategic-electoral construction of RN was strongest in 2022–2024 and can be interrupted by specific legal and political events.

The final majority-vote analysis disciplines claims that might otherwise be overstated: the pooled blame and target coefficients are near zero or non-significant (\texttt{blame\_frame} adjusted RN OR = 0.966, \emph{p} = 0.426; \texttt{actor\_role} = TARGET adjusted RN OR = 1.049, ns) and should not be reported as simple aggregate party differences.

But they are not empty nulls; they are structural aggregation nulls: the pooled coefficient cancels left- and right-leaning outlet logics of comparable magnitude pointing in opposite directions. The outlet-level interaction structure (below) carries the substantive signal. The blame null in particular should not be read as balanced blame attribution between the parties; blame attribution may also be harder to detect reliably from headlines alone, where only the surface framing act is observable and the attributed moral responsibility is often implicit.

Normalization is RN-positive but slight (43.7\% LFI vs. 45.1\% RN; RN OR = 1.080, $p < 0.01$) and is a secondary finding. The personalization row runs opposite: leader-centric or mixed headlines are more common for LFI (51.4\% vs. 47.7\%; RN OR = 0.907, $p < 0.001$), consistent with LFI coverage routed more through Mélenchon and other named leaders; because personalization is not a framing register, it is interpretive context, not a central claim.

\begin{table}[t]
\centering
\scriptsize
\setlength{\tabcolsep}{3pt}
\renewcommand{\arraystretch}{1.03}
\resizebox{\columnwidth}{!}{%
\begin{tabular}{@{}llccc@{}}
\toprule
Outlet & Orient. & Delegit & Target & Blame \\
\midrule
Libération & C-left & 3.646*** & 2.043*** & 3.234*** \\
Le HuffPost & C-left & 2.216*** & 1.316* & 1.670* \\
Franceinfo & Public & 1.612*** & 1.366** & 0.954 ns \\
Le Monde & C-left & 1.741*** & 0.993 ns & 1.498 ns \\
Le Parisien & C/pop. & n/a & 1.209* & 1.242 ns \\
20 Minutes & Cent. & n/a & 1.199 ns & 1.197 ns \\
Le Figaro & C-right & 0.821* & 0.729*** & 0.587*** \\
JDD\textsuperscript{†} & Right & 0.485*** & 0.530*** & 0.284*** \\
Other & Mixed & n/a & 1.011 ns & 0.846** \\
\bottomrule
\end{tabular}
}
\caption{Outlet-level interaction ORs (RN vs.\ LFI). Orient.: C = centre, pop. = popular, Cent. = centrist. Delegit ORs: 6 outlets with sufficient cell size only (n/a = excluded). *** $p < 0.001$; ** $p < 0.01$; * $p < 0.05$; ns = not significant. \textsuperscript{†}JDD: Bolloré editorial shift, summer 2023 (§5, Appendix~\ref{app:corpus}).}
\label{tab:outlet-interactions}
\end{table}

\subsection{Behavioral Reality Versus Moral Accounting}

The sharpest structural contrast is between \texttt{aggressor} and the \texttt{target}/\texttt{blame} pair. The aggressor finding has a much weaker outlet interaction (LR(8) = 17.4, \emph{p} = 0.027) than target (LR(8) = 123.7, \emph{p} < 0.001) or blame (LR(8) = 129.8, \emph{p} < 0.001), which remain among the most outlet-structured findings in the dataset. This yields the paper's deepest result: broad agreement on who attacks, but sharp editorial disagreement on who suffers and who is responsible.
The per-model robustness analysis (Appendix Table~\ref{tab:per-model-robustness}) independently validates this two-tier structure: the behavioral tier survives in every model; the moral-accounting tier does not. Although the aggressor outlet interaction reaches formal significance, it is only about one-seventh the size of the target and blame interactions and lacks their left-right reversal, so the contrast remains one of shared behavioral narration versus polarized moral accounting.

The aggregate aggressor coefficient shows that LFI is more often cast as aggressor than RN (adjusted RN OR = 0.732, \emph{p} < 0.001), but AGGRESSOR is a behavioral and syntactic role (attacking, denouncing, pushing, threatening, or targeting), not a direct measure of hostile evaluation. When Libération writes that LFI "monte au créneau" [steps up to fight] or that "les insoumis ripostent" [the insoumis strike back], LFI is annotated as aggressor even when the outlet is covering LFI's activism approvingly. The finding is consistent with either LFI's genuinely combative political style or shared journalistic conventions for narrating left-wing activist politics; the data cannot adjudicate between them. The LFI aggressor excess is absent at Libération and JDD (ns), consistent with outlets narrating their preferred party's combativeness more symmetrically. Because actor-role reliability is lower than conflict and strategic-game reliability, we use this coefficient as corroborating syntax, not a primary measurement claim; the party-stratified audit shows nearly identical actor-role human--MV agreement for LFI (79.8\%) and RN (80.2\%), with no significant directional AGGRESSOR error (Appendix~\ref{tab:actor-party}).

The data show consistent LFI conflict excess across outlet families (Table~\ref{tab:outlet-interactions}) but diverge sharply on the moral consequence of that behavior: left outlets frame RN as the party under siege and bearing responsibility for political conflict, while right outlets frame LFI as unfairly targeted by a hostile establishment. This divergence is not confined to the clearest ideological poles: beyond \emph{Libération} and \emph{JDD}, RN-positive target tilts also appear at \emph{Le HuffPost}, \emph{Le Parisien}, and \emph{Franceinfo} (post-hoc outlet-level tests, not corrected for multiple comparisons). Frame co-occurrence analysis (Appendix~\ref{app:cooccurrence}) corroborates the two-tier structure: \texttt{target} × delegitimization is strongest ($\varphi = +0.684$), followed by \texttt{blame} × delegitimization ($\varphi = +0.474$), \texttt{conflict} × \texttt{aggressor} ($\varphi = +0.473$), and \texttt{strategic\_game} × normalization ($\varphi = +0.440$); delegitimization and normalization are strongly negative ($\varphi = -0.566$). Figure~\ref{fig:forest} summarizes the adjusted party odds ratios across all three evidentiary tiers.

\section{Discussion}

An accurate-coverage explanation remains plausible: LFI, under Mélenchon and then Panot, led Article~49.3 opposition, censure motions, and street mobilizations against pension reform, while the FN/RN under Marine Le Pen pursued a dédiabolisation strategy aimed at normalization and broader electoral appeal \citep{ivaldi2014}. The observational design cannot distinguish behavioral supply from framing construction.

For delegitimization, the JDD editorial transition is the most suggestive within-outlet discontinuity: after Bolloré's summer-2023 consolidation of editorial control, JDD's delegitimization direction reversed (Table~\ref{tab:outlet-interactions} and Appendix~\ref{app:corpus}). Because there is no matched control outlet, the shift coincides with the high-conflict pension-reform period, and the pre-Bolloré window is thin, we treat this as suggestive rather than quasi-experimental.

French news headlines assign LFI and RN \emph{different political roles}, with outlet-level patterns organizing the moral-accounting layer around a behavioral core shared across outlets. That asymmetry makes this a study of \emph{editorial differentiation} rather than aggregate partisanship: generic sentiment or negativity measures cannot distinguish a press system that dislikes one party from one casting two parties as different \emph{kinds} of political object. Post-hoc sentiment supports this: adjusted sentiment ORs are statistically significant but substantively small (0.716--0.925), while conflict-frame ORs are larger and direction-stable (0.533--0.688; Table~\ref{tab:sentiment-robustness}).

\section{Limitations}

\paragraph{LLM annotator dependence and correlated biases.}
The three annotation models (GPT-OSS-120B, Llama-3.3-70B-Instruct, and Mistral Large 2 (123B)) are not statistically independent annotators. All three are trained on large web-scale corpora that include French-language political news from the study period; despite different providers and architectural families, shared exposure to similar distributional patterns in French political text cannot be ruled out. Majority vote reduces random disagreement but not shared systematic bias: if all three models have internalized the same press framing they measure, estimates of role-assignment frequency could be inflated without any single model being visibly wrong.

Several features of the evidence argue against circularity as the \emph{primary} explanation for cross-model agreement. First, the two-tier reliability pattern is inconsistent with pure training-data replication: if models were uniformly recovering encoded press frames, agreement rates should be high across fields; instead, the behavioral tier achieves 79.6--85.5\% unanimous agreement while the moral-accounting tier achieves only 53.4\%, with Mistral Large 2 reversing direction on delegitimization entirely (OR = 0.984, ns) while GPT-OSS and Llama agree. Cross-provider disagreement of this magnitude on a specific dimension is more consistent with genuine annotation difficulty than with models echoing a shared representation. Second, the 400-headline human validation audit (coded blind to model output) corroborates the behavioral findings above the pre-specified threshold ($\kappa \geq 0.76$ for conflict and strategic-game), providing an external anchor not subject to LLM training-data confounds. Third, the behavioral-tier asymmetry holds in centrist outlets (Franceinfo, Le Parisien, 20 Minutes) unlikely to systematically over-represent LFI conflict framing in their training data.

To bound this empirically, we computed cross-model three-way unanimous agreement rates by publication year across 28,592 shared headlines (GPT-OSS-120B, Llama-3.3-70B-Instruct, Mistral Large 2). If training-data circularity were the primary driver, 2025 headlines (most likely to fall outside or near the training-data boundary of at least some models) should show materially lower agreement. They do not. For \texttt{conflict\_frame}, 2025 unanimous agreement is 84.2\% versus a 2022--2024 baseline of 85.4\% ($\Delta = -1.2\,\mathrm{pp}$). For \texttt{strategic\_game\_frame}, 2025 agreement is 84.3\% versus a baseline of 79.0\% ($\Delta = +5.3\,\mathrm{pp}$). Across both behavioral-tier constructs, 2025 agreement is within the year-to-year variation observed over 2022--2024 and shows no systematic decline. This pattern does not show the decline a simple cutoff-based circularity account would predict: if models were merely recovering memorized press framings rather than detecting real signal, the 2025 subset, least likely to fall inside every model's training window, would be the natural place for agreement to drop, and it does not. We do not read this as ruling circularity out: stable agreement is also consistent with year-invariant distributional regularities shared across models, or with partial 2025 exposure for some models. It does, however, remove the most cutoff-specific version of the concern. Residual exposure to 2025 training data cannot be ruled out for any of the three models, and adversarial prompting or calibration against a fully human-annotated gold standard remains a natural extension. However, the survival of the behavioral asymmetry across ideologically centrist outlets and blinded human coders provides an external anchor, showing that the pipeline detects structured editorial differentiation rather than merely echoing memorized LLM training bias.

\paragraph{Legitimacy field reliability.}
The legitimacy dimension (whether a headline delegitimizes, legitimizes, or is neutral toward the party) is the weakest field in the human-versus-majority validation study, with lower inter-annotator agreement in human coding and the largest gap between majority-vote and human labels. Aggregate patterns on the legitimacy dimension (e.g., outlet-ideology effects on delegitimization) are therefore more uncertain than the role-assignment findings, and case-level legitimacy labels should not be treated as reliable for individual headlines. The moral-accounting layer reported in the results section is supported by aggregate patterns across thousands of headlines, but individual-label uncertainty warrants caution in any downstream application that depends on per-headline delegitimization scores.

\paragraph{Keyword retrieval and syntactic sampling.}
The corpus was constructed by retrieving headlines containing party-name keywords. This procedure necessarily oversamples headlines in which the party or its leaders are mentioned by name, and may oversample headlines in which the party appears in subject position, the syntactic slot most strongly associated with agency and conflict framing. A position-proxy check on 100 randomly sampled headlines per party finds RN keywords in the first-third of the headline in 43.4\% of cases versus 34.5\% for LFI ($z = -1.25$, $p = 0.21$), providing no significant evidence of differential subject-position retrieval. However, the check is approximate: keyword position is not equivalent to grammatical subject position, and the sample is small. The possibility that LFI appears more frequently as syntactic agent in the full corpus, independently of keyword position, cannot be ruled out with the current design. Replication with a full-text dependency parse on a random subsample is a natural robustness extension.

\paragraph{Observational design and causal scope.}
The study is observational: it documents asymmetric role assignment in the corpus as it exists, but cannot identify the mechanisms that produce it. Three classes of explanation are consistent with the pattern: behavioral reality (LFI actually generates more political conflict events), journalistic convention (conflict framing is an established editorial template applied selectively), and editorial ideology (outlets strategically differentiate challengers). The design cannot adjudicate between them. This limitation is particularly consequential for the conflict asymmetry: LFI was the primary architect of France's most conflict-intensive parliamentary episodes between 2022 and 2025 (pension-reform opposition under Article 49.3, repeated censure motions, street mobilization), while RN was simultaneously executing a deliberate normalization strategy. The observed headline pattern is therefore equally consistent with accurate coverage of genuinely different political behaviors and with framing construction. A first-pass empirical check, correlating annual headline conflict rates with per-quarter counts of parliamentary interventions per party from the Assemblée nationale's public record, is a feasible future test that the current design cannot provide. The temporal and outlet-stratified analyses provide circumstantial evidence (role asymmetry holds in centrist outlets; it predates and postdates LFI's period of peak parliamentary conflict), but these are consistency checks, not causal identification. Designs that could disentangle behavioral supply from framing demand (e.g., matching headline framing to event-level political-conflict databases such as parliamentary speech counts, censure motions, or recorded demonstrations, or exploiting outlet-level editorial shifts as quasi-experiments) are the natural next step for this research program.

\paragraph{Generalizability.}
The corpus covers 25 French-language outlets in metropolitan France between 2022 and 2025. Generalization to other national press systems, other time windows, or other left/right challenger pairs requires separate empirical work. French political journalism is characterized by high outlet pluralism, a historically rooted suspicion of populist challengers from both poles, and a specific post-2017 party-system configuration (LFI and RN as the two dominant challengers to a centrist presidential coalition) that may not translate directly to other cases. The annotation scheme was developed and validated on LFI and RN; whether the same scheme and agreement thresholds generalize to governing parties or smaller centrist actors (EELV, LR, En Marche) is an open empirical question that the current design cannot answer. The finding that role asymmetry survives across outlet families and across most of 2022--2025 provides internal robustness, but does not establish that the pattern is invariant across press cultures or party types.

\paragraph{Headline-only scope.}
All analysis is restricted to headlines. Headlines are high-salience framing units that reach audiences who never read the article body \citep{gabielkov2016}, but they are not representative of full article content. Headline framing can diverge from the framing of the article body \citep{ecker2014}. Extending the annotation scheme to lead paragraphs and article bodies is a natural next step, but would require a substantially larger human validation study to establish comparably reliable label quality.

\section{Ethics Statement}

\paragraph{Data and privacy.}
The corpus consists entirely of publicly published news headlines retrieved from French-language media outlets. No personal data were collected. No social media data, user data, or private communications are included. The analysis is at the level of editorial output (published headlines and outlet-level aggregate patterns), not at the level of individual journalists, editors, or readers. All outlets are publicly named in the supplementary material; none is anonymized, as outlet identity is a variable in the analysis.

\paragraph{Human annotation.}
The human validation study involved two annotators who coded 400 headlines. Annotators were informed of the research purpose, including that the study examines asymmetric political framing of LFI and RN. They were compensated at or above the applicable minimum hourly rate for research annotation work. No personally identifying information about annotators is reported. The annotation task did not involve exposure to harmful, violent, or distressing content; the headlines are standard political journalism.

\paragraph{LLM-based annotation and potential biases.}
We use large language models as primary annotators. LLMs trained on web-scale corpora may have internalized existing political biases present in French-language text, including the very framing asymmetries this paper documents. Results should be interpreted with this circularity risk in mind: the models may be replicating rather than independently detecting press-level patterns. We report inter-model agreement and human-validation accuracy explicitly to allow readers to assess the degree to which the annotation pipeline introduces systematic distortion. We do not claim that majority-vote LLM annotation is equivalent to independent human annotation; the human audit is included precisely because it is not.

\paragraph{Political framing research and potential misuse.}
This paper documents asymmetric political role assignment in French news coverage. The findings could in principle be used to support narratives that one party receives unfair press treatment. We note that the paper does not evaluate the accuracy, fairness, or appropriateness of the framing patterns it documents; it measures what roles headlines assign, not whether those roles are warranted by the parties' actual behavior. Readers and downstream users should not interpret role-assignment frequencies as editorial verdicts. The study is designed to advance the methodology of computational framing research and to contribute to comparative political communication, not to adjudicate editorial fairness claims about specific outlets or parties.

\paragraph{Reproducibility.}
The annotation prompts, validation materials, and aggregate corpus statistics necessary to reproduce the main findings are included in the supplementary material. Model versions are reported in the supplementary material to support reproducibility given that model behavior may change across API versions. The supplementary package is provided for review purposes; post-publication data availability will be determined upon acceptance.

\bibliography{paper_refs}

\begin{thebibliography}{27}
\providecommand{\natexlab}[1]{#1}

\bibitem[{Benson and Hallin(2007)}]{bensonHallin2007}
Rodney Benson and Daniel~C. Hallin. 2007.
\newblock \href {https://doi.org/10.1177/0267323107073746} {How states, markets and globalization shape the news: The french and us national press, 1965--97}.
\newblock \emph{European Journal of Communication}, 22(1):27--48.

\bibitem[{Cag{\'e} et~al.(2022)Cag{\'e}, Hengel, Herv{\'e}, and Urvoy}]{cage2022}
Julia Cag{\'e}, Moritz Hengel, Nicolas Herv{\'e}, and Camille Urvoy. 2022.
\newblock \href {https://doi.org/10.2139/ssrn.4036211} {Hosting media bias: Evidence from the universe of french broadcasts, 2002--2020}.
\newblock Technical Report hal-03878119, Sciences Po / HAL.

\bibitem[{Card et~al.(2015)Card, Boydstun, Gross, Resnik, and Smith}]{card2015}
Dallas Card, Amber~E. Boydstun, Justin~H. Gross, Philip Resnik, and Noah~A. Smith. 2015.
\newblock \href {https://doi.org/10.3115/v1/P15-2072} {The media frames corpus: Annotations of frames across issues}.
\newblock In \emph{Proceedings of the 53rd Annual Meeting of the Association for Computational Linguistics and the 7th International Joint Conference on Natural Language Processing (Volume 2: Short Papers)}, pages 438--444, Beijing, China. Association for Computational Linguistics.

\bibitem[{{CEVIPOF}(2026)}]{cevipof2019barometre}
{CEVIPOF}. 2026.
\newblock \href {https://www.sciencespo.fr/cevipof/barometre-confiance-politique} {Barom{\`e}tre de la confiance politique - vague 17}.
\newblock Sciences Po CEVIPOF.

\bibitem[{Ecker et~al.(2014)Ecker, Lewandowsky, Chang, and Pillai}]{ecker2014}
Ullrich K.~H. Ecker, Stephan Lewandowsky, Ee~Pin Chang, and Rekha Pillai. 2014.
\newblock \href {https://doi.org/10.1037/xap0000028} {The effects of subtle misinformation in news headlines}.
\newblock \emph{Journal of Experimental Psychology: Applied}, 20(4):323--335.

\bibitem[{Entman(1993)}]{entman1993}
Robert~M. Entman. 1993.
\newblock \href {https://doi.org/10.1111/j.1460-2466.1993.tb01304.x} {Framing: Toward clarification of a fractured paradigm}.
\newblock \emph{Journal of Communication}, 43(4):51--58.

\bibitem[{Fan et~al.(2019)Fan, White, Sharma, Su, Choubey, Huang, and Wang}]{fan2019}
Lisa Fan, Marshall White, Eva Sharma, Ruisi Su, Prafulla~Kumar Choubey, Ruihong Huang, and Lu~Wang. 2019.
\newblock \href {https://doi.org/10.18653/v1/D19-1664} {In plain sight: Media bias through the lens of factual reporting}.
\newblock In \emph{Proceedings of the 2019 Conference on Empirical Methods in Natural Language Processing and the 9th International Joint Conference on Natural Language Processing (EMNLP-IJCNLP)}, pages 4771--4781.

\bibitem[{Gabielkov et~al.(2016)Gabielkov, Ramachandran, Chaintreau, and Legout}]{gabielkov2016}
Maksym Gabielkov, Arthi Ramachandran, Augustin Chaintreau, and Arnaud Legout. 2016.
\newblock \href {https://doi.org/10.1145/2896377.2901462} {Social clicks: What and who gets read on twitter?}
\newblock In \emph{ACM SIGMETRICS / IFIP Performance 2016}, Antibes Juan-les-Pins, France.

\bibitem[{Gamson and Modigliani(1989)}]{gamsonModigliani1989}
William~A. Gamson and Andre Modigliani. 1989.
\newblock \href {https://doi.org/10.1086/229213} {Media discourse and public opinion on nuclear power: A constructionist approach}.
\newblock \emph{American Journal of Sociology}, 95(1):1--37.

\bibitem[{Graves-Sandriman(2026)}]{gravesSandriman2026}
Emma~Kaylee Graves-Sandriman. 2026.
\newblock \href {https://doi.org/10.65476/82q8mw89} {A model of frame categories for analyzing media discourse of emerging technologies}.
\newblock \emph{International Journal of Communication}, 20.

\bibitem[{Heseltine and Clemm~von Hohenberg(2024)}]{heseltine2024}
Michael Heseltine and Bernhard Clemm~von Hohenberg. 2024.
\newblock \href {https://doi.org/10.1177/20531680241236239} {Large language models as a substitute for human experts in annotating political text}.
\newblock \emph{Research \& Politics}, 11(1).

\bibitem[{Ivaldi(2014)}]{ivaldi2014}
Gilles Ivaldi. 2014.
\newblock \href {https://shs.hal.science/halshs-01059581v1} {A new course for the french radical-right? the {Front National} and dedemonization}.
\newblock Seminar ``Up to the Mainstream? Radical Right Parties in Western Europe. 2000--2013'', University of Amsterdam, 19--20 June 2014. HAL Id: halshs-01059581.

\bibitem[{Landis and Koch(1977)}]{landisKoch1977}
J.~Richard Landis and Gary~G. Koch. 1977.
\newblock \href {https://doi.org/10.2307/2529310} {The measurement of observer agreement for categorical data}.
\newblock \emph{Biometrics}, 33(1):159--174.

\bibitem[{Mahmoud et~al.(2025)Mahmoud, Xie, Dimitrov, Nikolaidis, Silvano, Yangarber, Sharma, Sartori, Stefanovitch, Da~San~Martino, Piskorski, and Nakov}]{mahmoud2025entity}
Tarek Mahmoud, Zhuohan Xie, Dimitar Dimitrov, Nikolaos Nikolaidis, Purifica{\c c}{\~a}o Silvano, Roman Yangarber, Shivam Sharma, Elisa Sartori, Nicolas Stefanovitch, Giovanni Da~San~Martino, Jakub Piskorski, and Preslav Nakov. 2025.
\newblock \href {https://doi.org/10.18653/v1/2025.findings-acl.17} {Entity framing and role portrayal in the news}.
\newblock In \emph{Findings of the Association for Computational Linguistics: ACL 2025}, pages 302--326, Vienna, Austria. Association for Computational Linguistics.

\bibitem[{McCombs and Shaw(1972)}]{mccombsShaw1972}
Maxwell~E. McCombs and Donald~L. Shaw. 1972.
\newblock \href {https://doi.org/10.1086/267990} {The agenda-setting function of mass media}.
\newblock \emph{Public Opinion Quarterly}, 36(2):176--187.

\bibitem[{{Meta AI}(2024)}]{meta2024llama33}
{Meta AI}. 2024.
\newblock Llama 3.3 70{B} instruct.
\newblock \url{https://huggingface.co/meta-llama/Llama-3.3-70B-Instruct}.
\newblock Accessed: 2026-05-26.

\bibitem[{{Mistral AI}(2024)}]{mistral2024large2}
{Mistral AI}. 2024.
\newblock Mistral large 2.
\newblock \url{https://huggingface.co/mistralai/Mistral-Large-Instruct-2407}.
\newblock Accessed: 2026-05-26.

\bibitem[{Mudde and Rovira~Kaltwasser(2017)}]{mudde2017}
Cas Mudde and Crist{\'o}bal Rovira~Kaltwasser. 2017.
\newblock \href {https://doi.org/10.1093/actrade/9780190234874.001.0001} {\emph{Populism: A Very Short Introduction}}.
\newblock Oxford University Press.

\bibitem[{Neuendorf(2017)}]{neuendorf2017}
Kimberly~A. Neuendorf. 2017.
\newblock \emph{The Content Analysis Guidebook}, 2 edition.
\newblock SAGE Publications.

\bibitem[{{OpenAI}(2025)}]{openai2025gptoss}
{OpenAI}. 2025.
\newblock \href {https://doi.org/10.48550/arXiv.2508.10925} {{gpt-oss-120b \& gpt-oss-20b Model Card}}.
\newblock Technical report, OpenAI.
\newblock ArXiv:2508.10925.

\bibitem[{Otmakhova et~al.(2024)Otmakhova, Khanehzar, and Frermann}]{otmakhova2024}
Yulia Otmakhova, Shima Khanehzar, and Lea Frermann. 2024.
\newblock \href {https://doi.org/10.18653/v1/2024.acl-long.822} {Media framing: A typology and survey of computational approaches across disciplines}.
\newblock In \emph{Proceedings of the 62nd Annual Meeting of the Association for Computational Linguistics (Volume 1: Long Papers)}, pages 15407--15428, Bangkok, Thailand. Association for Computational Linguistics.
\newblock Outstanding Paper Award, ACL 2024.

\bibitem[{Piskorski et~al.(2023)Piskorski, Stefanovitch, {Da San Martino}, and Nakov}]{piskorski2023}
Jakub Piskorski, Nicolas Stefanovitch, Giovanni {Da San Martino}, and Preslav Nakov. 2023.
\newblock \href {https://doi.org/10.18653/v1/2023.semeval-1.317} {{SemEval}-2023 task 3: Detecting the category, the framing, and the persuasion techniques in online news in a multi-lingual setup}.
\newblock In \emph{Proceedings of the 17th International Workshop on Semantic Evaluation ({SemEval}-2023)}, pages 2343--2361.

\bibitem[{Semetko and Valkenburg(2000)}]{semetko2000}
Holli~A. Semetko and Patti~M. Valkenburg. 2000.
\newblock \href {https://doi.org/10.1111/j.1460-2466.2000.tb02843.x} {Framing european politics: A content analysis of press and television news}.
\newblock \emph{Journal of Communication}, 50(2):93--109.

\bibitem[{Shoemaker and Vos(2009)}]{shoemakerVos2009}
Pamela~J. Shoemaker and Tim~P. Vos. 2009.
\newblock \emph{Gatekeeping Theory}.
\newblock Routledge, New York.

\bibitem[{Spinde et~al.(2021)Spinde, Plank, Krieger, Ruas, Gipp, and Aizawa}]{spinde2021}
Timo Spinde, Manuel Plank, Jan-David Krieger, Terry Ruas, Bela Gipp, and Akiko Aizawa. 2021.
\newblock \href {https://doi.org/10.18653/v1/2021.findings-emnlp.101} {Neural media bias detection using distant supervision with babe - bias annotations by experts}.
\newblock In \emph{Findings of the Association for Computational Linguistics: EMNLP 2021}, pages 1166--1177, Punta Cana, Dominican Republic. Association for Computational Linguistics.

\bibitem[{T{\"o}rnberg(2025)}]{tornberg2025}
Petter T{\"o}rnberg. 2025.
\newblock \href {https://doi.org/10.1177/08944393241286471} {Large language models outperform expert coders and supervised classifiers at annotating political social media messages}.
\newblock \emph{Social Science Computer Review}, 43(6):1181--1195.

\bibitem[{Vallejo~Vera and Driggers(2025)}]{vallejo2025}
Sebasti{\'a}n Vallejo~Vera and Hunter Driggers. 2025.
\newblock \href {https://doi.org/10.1057/s41599-025-05834-4} {{LLMs} as annotators: The effect of party cues on labelling decisions by large language models}.
\newblock \emph{Humanities and Social Sciences Communications}, 12:1530.

\end{thebibliography}

\clearpage
\appendix

\makeatletter
\@addtoreset{table}{section}
\@addtoreset{figure}{section}
\makeatother
\setcounter{table}{0}
\setcounter{figure}{0}
\renewcommand{\thetable}{\thesection.\arabic{table}}
\renewcommand{\thefigure}{\thesection.\arabic{figure}}

\section{Corpus Composition and Retrieval Scope}
\label{app:corpus}

\begin{table*}[t]
\centering
\scriptsize
\setlength{\tabcolsep}{4pt}
\renewcommand{\arraystretch}{1.05}
\begin{tabularx}{\textwidth}{@{}>{\RaggedRight\arraybackslash}X>{\RaggedRight\arraybackslash}Xrrrr@{}}
\toprule
Outlet & Orientation & LFI & RN & BOTH & Total \\
\midrule
Le Figaro & Centre-right / right & 1,298 & 1,290 & 106 & 2,694 \\
Franceinfo & Public / centrist & 1,012 & 1,556 & 81 & 2,649 \\
Le Parisien & Centre / popular & 1,092 & 1,233 & 119 & 2,444 \\
Libération & Centre-left & 905 & 1,087 & 54 & 2,046 \\
JDD\textsuperscript{†} & Centre-right (pre-2023) / Right (post-2023) & 1,006 & 811 & 71 & 1,888 \\
BFMTV & Centre / TV news & 822 & 844 & 73 & 1,739 \\
CNEWS & Right / conservative & 727 & 806 & 81 & 1,614 \\
Le Point & Centre-right & 885 & 619 & 83 & 1,587 \\
20 Minutes & Centrist / free press & 660 & 808 & 51 & 1,519 \\
Le HuffPost & Centre-left & 721 & 625 & 54 & 1,400 \\
Le Monde & Centre-left & 472 & 872 & 40 & 1,384 \\
Valeurs actuelles & Far-right & 432 & 565 & 36 & 1,033 \\
Le Nouvel Obs & Centre-left & 419 & 490 & 29 & 938 \\
TF1 INFO & Centre / TV news & 329 & 505 & 36 & 870 \\
La Croix & Christian-democrat & 322 & 452 & 28 & 802 \\
Fdesouche & Far-right aggregator & 371 & 274 & 45 & 690 \\
L'Humanité & Left / communist & 235 & 385 & 12 & 632 \\
Les Echos & Liberal / business & 188 & 404 & 15 & 607 \\
Marianne & Sovereignist / republican & 282 & 293 & 28 & 603 \\
L'Express & Liberal / centre-right & 256 & 313 & 25 & 594 \\
Mediapart & Investigative / left & 136 & 310 & 5 & 451 \\
Ouest-France & Regional / centrist & 70 & 126 & 6 & 202 \\
Causeur & Conservative / right & 69 & 55 & 0 & 124 \\
Slate.fr & Digital / centrist & 30 & 27 & 3 & 60 \\
Blast & Digital / left & 2 & 20 & 0 & 22 \\
\textbf{Total} & & \textbf{12,741} & \textbf{14,770} & \textbf{1,081} & \textbf{28,592} \\
\bottomrule
\end{tabularx}
\caption{Corpus composition by outlet.\label{tab:appendix-corpus}}
\end{table*}

Orientation labels are descriptive assignments based on established French media typologies; they are not used as ground-truth scores in the models. BOTH = headlines jointly framing LFI and RN. \textsuperscript{†}JDD shifted from centre-right to right under Bolloré ownership in summer 2023. A pre/post-2023 split confirms the direction reversal: delegitimization rate-ratio was 1.52 (RN\,>\,LFI) pre-shift and 0.54 (LFI\,>\,RN) post-shift; the post-Bolloré stance accounts for 83\% of JDD's corpus rows.

The retrieval scope included LFI, RN, LFI-led coalitions (NUPES, NFP), and high-salience leaders (Mélenchon, Panot, Aubry; Le Pen, Bardella). Coalition headlines were retained only when the majority-vote \texttt{party\_target} label confirmed LFI as the primary target: 1,666 of 2,130 coalition candidates (78.2\%) were retained; the remainder were classified as BOTH, RN, or UNCLEAR and excluded.

Final retrieval rules used the following party-specific terms. LFI-side terms were \texttt{LFI}, \texttt{La France Insoumise}, \texttt{France Insoumise}, \texttt{insoumis/insoumise}, \texttt{NUPES}, \texttt{NFP}, \texttt{Nouveau Front Populaire}, \texttt{Mélenchon}, \texttt{Mathilde Panot}, and \texttt{Manon Aubry}. RN-side terms were \texttt{RN}, \texttt{Rassemblement National}, \texttt{Marine Le Pen}, and \texttt{Jordan Bardella}.

Excluded terms: secondary LFI deputies (e.g., Bompard; omitted to preserve symmetry with the RN side), Bruno Retailleau (not an RN actor), bare \texttt{Le Pen}/\texttt{Bardella}, Jean-Marie Le Pen, Maréchal, Philippot, Bay, Ciotti, and generic \texttt{extrême droite} references.

\section{Human Validation and Validation Rules}
\label{app:human-validation}

Three strata: disagreement-heavy (\emph{n}=150), high-agreement controls (\emph{n}=150), theory-targeted oversamples (\emph{n}=100). Split rows are overrepresented; stratum rates are not corpus-level accuracy estimates. Coder disagreements were reconciled via a codebook-only adjudication pass, blind to model labels.

\begin{table*}[t]
\centering
\small
\begin{tabular*}{\textwidth}{@{\extracolsep{\fill}}lrrrrl@{}}
\toprule
Field & vs. MV & vs. GPT & vs. Llama & vs. Mistral & MV best? \\
\midrule
\texttt{reference\_form} & 91.8 & \textbf{92.8} & 86.0 & 90.0 & no (GPT +1.0pp) \\
\texttt{blame\_frame} & 92.0 & 88.5 & \textbf{92.2} & 90.8 & no (Llama +0.2pp) \\
\texttt{strategic\_game\_frame} & \textbf{93.2} & 90.5 & 92.5 & 90.8 & yes \\
\texttt{conflict\_frame} & 89.8 & 84.0 & \textbf{90.2} & 88.2 & no (Llama +0.4pp) \\
\texttt{actor\_role} & \textbf{79.0} & 77.5 & 76.2 & 71.0 & yes \\
\texttt{legitimacy\_frame} & \textbf{74.5} & 69.8 & 68.0 & 71.5 & yes \\
\bottomrule
\end{tabular*}
\caption{Recomputed comparison between the frozen adjudicated human labels and the final GPT/Llama/Mistral ensemble. Because human labels were coded blind to model output, no new human coding was needed for this comparison.\label{tab:appendix-human-vs-model}}
\end{table*}

\begin{table*}[t]
\centering
\scriptsize
\begin{tabular*}{\textwidth}{@{\extracolsep{\fill}}lrrrrrr@{}}
\toprule
Field & Unanimous n & H-MV & 2-of-3 n & H-MV & Split n & H-MV \\
\midrule
\texttt{reference} & 326 & 98.5 & 65 & 66.2 & 9 & 33.3 \\
\texttt{blame} & 348 & 95.7 & 52 & 67.3 & n/a & n/a \\
\texttt{strategic} & 348 & 96.3 & 52 & 73.1 & n/a & n/a \\
\texttt{conflict} & 329 & 94.2 & 71 & 69.0 & n/a & n/a \\
\texttt{actor} & 246 & 91.9 & 134 & 58.2 & 20 & 60.0 \\
\texttt{legitimacy} & 191 & 92.7 & 193 & 60.1 & 16 & 31.2 \\
\bottomrule
\end{tabular*}
\caption{Human-majority agreement by final GPT/Llama/Mistral model-agreement stratum. H-MV denotes agreement between the adjudicated human label and the majority-vote label.\label{tab:appendix-human-strata}}
\end{table*}

Pre-registered rules: human-majority agreement <70\% downgrades a field to sensitivity; split-stratum agreement <50\% restricts reliance on split rows. Legitimacy is caveated (weakest audited field, temporally/model sensitive, lowest unanimous-agreement rate, below threshold on three-way splits). Excluding three-way splits preserves the RN-positive direction (OR\,=\,1.258 [1.191, 1.330]); unanimous-only: OR\,=\,1.284 [1.195, 1.379]. We report delegitimization as a tendency, not a foreground claim, given its 2023 reversal and lower audit strength.

\paragraph{Party-stratified actor-role validation.}
\texttt{actor\_role} is validated more weakly than \texttt{conflict\_frame} and \texttt{strategic\_game\_frame} (human $\kappa = 0.614$ vs.\ $0.762$/$0.805$; Table~\ref{tab:reliability}), and AGGRESSOR is the most politically loaded label in the scheme. We therefore test whether that weaker measurement is \emph{asymmetric} between LFI and RN, since a party-directional actor-role error would threaten the headline conflict/aggressor result, whereas a uniform attenuation would not. Using the frozen blind audit (gold = adjudicated human label; prediction = canonical majority vote), human--MV \texttt{actor\_role} agreement is statistically indistinguishable across parties: LFI 79.8\% ($\kappa = 0.696$, \emph{n} = 168), RN 80.2\% ($\kappa = 0.660$, \emph{n} = 187), a 0.4\,pp gap (two-proportion $z = -0.11$, $p = 0.92$); the joint BOTH subset is lower (71.1\%, \emph{n} = 45) and is excluded from the main contrast in any case. Restricting to the AGGRESSOR label, the majority vote is \emph{conservative} rather than party-biased: it under-detects AGGRESSOR symmetrically (recall 0.58 LFI / 0.48 RN; precision 0.88 LFI / 1.00 RN), and the false-AGGRESSOR rate is near zero for both parties (LFI 2.3\%, \emph{n} = 132; RN 0.0\%, \emph{n} = 164). Neither directional error rate differs significantly by party (missed-AGGRESSOR gap $-10.5$\,pp, $p = 0.43$; false-AGGRESSOR gap $+2.3$\,pp, $p = 0.052$; Table~\ref{tab:actor-party}). The weaker actor-role reliability is thus a uniform attenuation toward the null that, if anything, biases \emph{against} the observed LFI aggressor excess rather than manufacturing it. Accordingly we present aggressor as corroborating role syntax rather than a primary measurement claim. These numbers are regenerated by the standalone verification module \texttt{verification/run\_all.py --module actor\_role\_party\_stratified}, which hard-checks reproduction of the authoritative overall 316/400 = 79.0\% before reporting any split.

\begin{table*}[t]
\centering
\footnotesize
\setlength{\tabcolsep}{4pt}
\renewcommand{\arraystretch}{1.08}
\begin{tabular*}{\textwidth}{@{\extracolsep{\fill}}lrrrrrr@{}}
\toprule
Party & \emph{n} & H--MV \% & $\kappa$ & AGG recall & AGG prec. & False-AGG \% \\
\midrule
LFI & 168 & 79.8 & 0.696 & 0.58 & 0.88 & 2.3 \\
RN & 187 & 80.2 & 0.660 & 0.48 & 1.00 & 0.0 \\
\midrule
\multicolumn{7}{@{}l}{\textit{LFI vs.\ RN: agreement gap 0.4\,pp ($z=-0.11$, $p=0.92$);}}\\
\multicolumn{7}{@{}l}{\textit{missed-AGG gap $p=0.43$; false-AGG gap $p=0.052$; symmetric.}}\\
\bottomrule
\end{tabular*}
\caption{Party-stratified \texttt{actor\_role} human--MV validation on the frozen blind audit. AGG = AGGRESSOR label. Errors are symmetric across parties and conservative (under-detection), so aggressor is retained as corroborating role syntax, not a primary claim.\label{tab:actor-party}}
\end{table*}

\section{Annotation Prompt Excerpt}
\label{app:prompt-excerpt}

Below is a condensed excerpt of the annotation prompt used for all three models (full prompt: \texttt{prompts/lfi\_rn\_prompt\_v5\_3.py} in the code repository). The system prompt instructed models to output valid JSON only, with exactly the nine fields defined in Table~\ref{tab:annotation-scheme}.

\paragraph{Field definitions as delivered to annotators.}

\begin{itemize}[leftmargin=1.2em]
\item \texttt{conflict\_frame}: TRUE when the headline features an \emph{explicit confrontation verb} between named political actors (e.g., \emph{attaque, dénonce, cible, torpille, s'en prend à}). FALSE when the headline reports an emotional reaction, attributed quote, analytical commentary, or structural political situation without a direct confrontation verb.

\item \texttt{blame\_frame}: TRUE when the party/leader is \emph{explicitly} presented as responsible for a concrete harm, wrongdoing, illegality, or morally harmful behavior (e.g., corruption, explicitly attributed hate speech). FALSE for: legal proceedings that merely report a trial exists without naming the wrongdoing; outsider rhetoric without specific wrongdoing attribution; prevention framing ("pour éviter que X arrive au pouvoir").

\item \texttt{strategic\_game\_frame}: TRUE when the headline frames politics as competition, tactical maneuver, or electoral game (electoral terms, parliamentary tactics, competitive framing, alliance arithmetic). FALSE for pure scandal or legal process without electoral emphasis.

\item \texttt{legitimacy\_frame}: NORMALIZING when the party is shown as an active participant in normal democratic life (any institutional action, electoral participation, or policy position). DELEGITIMIZING when the headline marks the party as deviant, dangerous, extremist, hypocritical, or subject to judicial sanction. NONE when the party appears as a passive object of description with no engagement signal.

\item \texttt{actor\_role}: AGGRESSOR when the party is the grammatical subject of a confrontation verb directed at a named opponent. TARGET when an identifiable agent explicitly attacks, judges, or condemns the party. NEUTRAL for institutional/electoral actions without an adversarial agent. MIXED when the party is genuinely both attacking and being attacked.
\end{itemize}

\paragraph{Worked boundary examples (conflict\_frame and legitimacy\_frame).}

\noindent\textit{conflict\_frame:} \textbf{TRUE}: ``LFI attaque le gouvernement sur les retraites'' [LFI attacks the government over pensions] (explicit confrontation verb + named opponents). \textbf{TRUE}: ``Le RN cible les insoumis pour leur vote sur le budget'' [RN targets the insoumis for their budget vote] (confrontation verb \emph{cible}; named target). \textbf{FALSE}: ``LFI ulcérée par les propos de X'' [LFI outraged by X's remarks] (emotional reaction; no confrontation verb). \textbf{FALSE}: ``Les propos du patron du RN font scandale'' [The RN leader's remarks cause a scandal] (scandal framing; no confrontation between named actors).

\noindent\textit{legitimacy\_frame:} \textbf{NORMALIZING}: ``Le RN dépose une motion de censure'' [RN files a motion of no confidence] (institutional parliamentary action). \textbf{NORMALIZING}: ``Marine Le Pen en tête dans les sondages pour 2027'' [Marine Le Pen leads in polls for 2027] (electoral participation signal). \textbf{DELEGITIMIZING}: ``Marine Le Pen condamnée à cinq ans d'inéligibilité'' [Marine Le Pen sentenced to five years of ineligibility] (judicial sanction). \textbf{DELEGITIMIZING}: ``Le RN aligné sur les positions de Moscou'' [RN aligned with Moscow's positions] (extremist/foreign-authoritarian framing). \textbf{NONE}: ``Le Rassemblement National présente son programme économique'' [Rassemblement National presents its economic programme] (party reports a position; no legitimacy signal in either direction).

\section{Codebook Boundary Rules}
\label{app:codebook}

Operational boundary rules (condensed; full prompt: \texttt{prompts/lfi\_rn\_prompt\_v5\_3.py}):

\begin{itemize}
\item \texttt{conflict\_frame}: true when the headline narrates attack, clash, denunciation, confrontation, threat, or explicit adversarial struggle.
\item \texttt{strategic\_game\_frame}: true when the headline foregrounds electoral competition, polling, coalition arithmetic, tactical positioning, campaign maneuver, or succession strategy.
\item \texttt{actor\_role}: AGGRESSOR when the party is syntactically doing the attacking or pressuring; TARGET when it is the object of attack or pressure; MIXED when both apply; NEUTRAL otherwise.
\item \texttt{legitimacy\_frame}: NORMALIZING when the party is treated as an ordinary democratic competitor; DELEGITIMIZING when the headline positions it as dangerous, scandalous, deviant, extremist, or outside acceptable democratic bounds; NONE when neither standing cue appears.
\item \texttt{blame\_frame}: true only when the headline attributes causal or moral responsibility for a concrete harm, failure, scandal, or crisis, not merely when it reports criticism.
\item \texttt{party\_target}: LFI, RN, BOTH, or UNCLEAR according to the primary party being framed; BOTH is used for joint or explicitly oppositional LFI/RN headlines.
\end{itemize}

\section{Temporal Lexical-Shift Diagnostics}
\label{app:lexical}

\begin{table*}[t]
\centering
\scriptsize
\setlength{\tabcolsep}{2pt}
\renewcommand{\arraystretch}{1.08}
\begin{tabular*}{\textwidth}{@{\extracolsep{\fill}}lrrrrrrrrrrr@{}}
\toprule
Year & \emph{n} LFI & \emph{n} RN & Conflict LFI & Conflict RN & Δ & Strategic LFI & Strategic RN & Δ & Delegit LFI & Delegit RN & Δ \\
\midrule
2022 & 3,729 & 3,733 & 0.103 & 0.086 & −0.017 & 0.494 & 0.559 & +0.065 & 0.173 & 0.207 & +0.034 \\
2023 & 2,502 & 2,020 & 0.175 & 0.104 & −0.071 & 0.193 & 0.343 & +0.150 & 0.321 & 0.288 & −0.033 \\
2024 & 3,815 & 5,375 & 0.123 & 0.081 & −0.042 & 0.422 & 0.526 & +0.104 & 0.245 & 0.334 & +0.089 \\
2025 & 2,695 & 3,642 & 0.174 & 0.087 & −0.087 & 0.300 & 0.322 & +0.022 & 0.329 & 0.400 & +0.071 \\
\bottomrule
\end{tabular*}
\caption{Year-level framing rates and LFI/RN gaps (2022–2025). Δ = RN rate minus LFI rate; raw proportions within year × party cells, not adjusted for outlet-composition differences across years.\label{tab:temporal}}
\end{table*}

\begin{figure}[t]
  \centering
  \includegraphics[width=0.82\columnwidth]{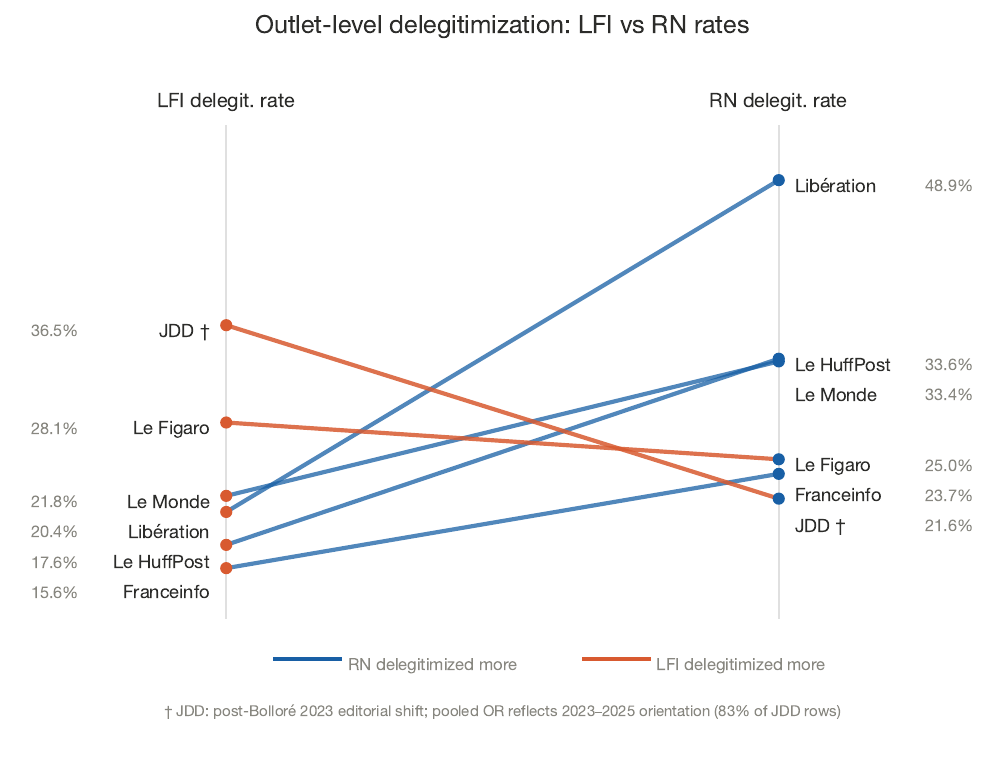}
  \caption{Outlet-level delegitimization rates for LFI (left) and RN (right). Lines slope up (blue) where RN is delegitimized more; lines slope down (coral) where LFI is delegitimized more. The JDD reversal reflects the post-Bol\-lo\-ré editorial shift; the aggregate OR of 1.291 averages over these opposing logics.\label{fig:outlet-slopes}}
\end{figure}

The lexical-shift diagnostic uses lowercased headline unigrams, removes common French stopwords and party-name tokens, and scores token distinctiveness with smoothed weighted log-odds z-scores. It is an interpretive aid, not a primary inferential test.

For each contrast, let $n_{1t}$ and $n_{2t}$ be counts of token $t$ in the focal and comparison headline groups, $N_1$ and $N_2$ the total retained-token counts in each group, and $\alpha_t=\max\{0.01(n_{1t}+n_{2t}),0.01\}$ the light empirical prior used in the script, with $\alpha_0=\sum_t \alpha_t$. Token distinctiveness is:

\[
\resizebox{\linewidth}{!}{$\displaystyle
\begin{aligned}
\delta_t &=
\log \frac{n_{1t}+\alpha_t}{N_1+\alpha_0-n_{1t}-\alpha_t}
-
\log \frac{n_{2t}+\alpha_t}{N_2+\alpha_0-n_{2t}-\alpha_t}, \\
z_t &= \frac{\delta_t}{\sqrt{\tfrac{1}{n_{1t}+\alpha_t}+\tfrac{1}{n_{2t}+\alpha_t}}}.
\end{aligned}
$}
\]

Positive $z_t$ = focal-group enriched; negative = comparison-group enriched. Tokens with fewer than eight total occurrences are excluded.

\begin{itemize}
\item LFI delegitimizing headlines in 2023 overrepresent \texttt{hamas}, \texttt{israel}, \texttt{retraites}, \texttt{reforme}, and \texttt{nupes}.
\item RN headlines in 2025 shift away from electoral terms such as \texttt{legislatives}, \texttt{europeennes}, \texttt{presidentielle}, and \texttt{majorite} toward judicial and ineligibility terms such as \texttt{condamnation}, \texttt{condamnee}, \texttt{ineligibilite}, \texttt{proces}, and \texttt{appel}.
\end{itemize}

\section{Frame Co-Occurrence Patterns}
\label{app:cooccurrence}

Table~\ref{tab:appendix-cooccurrence} reports the full co-occurrence structure underlying the two-tier interpretation discussed in the main text.

\begin{table*}[t]
\centering
\scriptsize
\begin{tabularx}{\textwidth}{@{}lrX@{}}
\toprule
Frame pair & $\phi$ & Interpretation \\
\midrule
\texttt{delegit} x \texttt{target} & +0.684 & Delegitimization and victimization co-occur strongly \\
\texttt{conflict} x \texttt{aggressor} & +0.473 & Conflict framing packages with aggressor role \\
\texttt{strategic\_game} x \texttt{normaliz} & +0.440 & Strategic frame co-occurs with normalization \\
\texttt{blame} x \texttt{delegit} & +0.474 & Blame and delegitimization are related but distinct \\
\texttt{delegit} x \texttt{normaliz} & -0.566 & Delegitimizing and normalizing frames are negatively associated \\
\texttt{target} x \texttt{normaliz} & -0.514 & Victim framing and normalization are negatively associated \\
\bottomrule
\end{tabularx}
\caption{Extended frame co-occurrence patterns.\label{tab:appendix-cooccurrence}}
\end{table*}

\section{Per-Model Robustness}
\label{app:per-model}

Table~\ref{tab:per-model-robustness} reports per-model estimates for all outcomes, showing that the behavioral-tier findings are direction-stable across models whereas the moral-accounting outcomes are not.

\begin{table*}[t]
\centering
\footnotesize
\setlength{\tabcolsep}{3.5pt}
\renewcommand{\arraystretch}{1.08}
\begin{tabular*}{\textwidth}{@{\extracolsep{\fill}}lrrrrl@{}}
\toprule
Outcome & MV & GPT-OSS & Llama & Mistral & Direction stable? \\
\midrule
\texttt{conflict\_frame} & 0.614*** & 0.688*** & 0.626*** & 0.533*** & Yes \\
\texttt{strategic\_game\_frame} & 1.414*** & 1.352*** & 1.643*** & 1.136*** & Yes \\
\texttt{aggressor} & 0.732*** & 0.833*** & 0.730*** & 0.656*** & Yes \\
\texttt{normaliz} & 1.080** & 1.014 ns & 1.186*** & 1.109*** & Yes \\
\texttt{leader} & 0.907*** & 0.914*** & 0.820*** & 0.889*** & Yes \\
\midrule
\texttt{delegit} & 1.291*** & 1.523*** & 1.317*** & 0.984 ns & No \\
\texttt{blame} & 0.966 ns & 1.038 ns & 0.931 ns & 1.070 ns & No \\
\texttt{target} & 1.048 ns & 1.065* & 1.107*** & 0.980 ns & No \\
\bottomrule
\end{tabular*}
\caption{Per-model odds ratios (RN vs.\ LFI) with outlet and year fixed effects. Direction-stable = all four sources agree on direction of effect. The behavioral tier (conflict, strategic-game, aggressor) is direction-stable and significant across all models independently. The moral-accounting tier (delegitimization, blame, target) is not: Mistral reverses direction on delegitimization and target.\label{tab:per-model-robustness}}
\end{table*}

\section{Cluster-Bootstrap Confidence Intervals}
\label{app:bootstrap}

Cluster-bootstrapped CIs (9 outlet families, 2,000 iterations; Table~\ref{tab:bootstrap-ci}) confirm the behavioral--moral-accounting distinction. Behavioral findings retain significance under outlet clustering; moral-accounting findings do not, consistent with the outlet interactions in §4.4. Wald intervals are anticonservative here because they treat within-outlet observations as independent.

All bootstrap models use the same logit specification (outcome $\sim$ party\_rn + C(outlet) + C(year)), with 9 outlet families: Le Figaro, Le Monde, Libération, Le Parisien, Le HuffPost, Franceinfo, JDD, 20 Minutes, and \emph{other} (17 remaining outlets pooled due to small cell sizes).

Nine clusters is below the $\geq$30 threshold standard guidance recommends for reliable asymptotic inference. For high intra-cluster-correlation outcomes (e.g., delegitimization, width ratio 7.17$\times$), intervals may be conservative. The behavioral-tier conclusions are unaffected.

\begin{table*}[t]
\centering
\footnotesize
\setlength{\tabcolsep}{3.5pt}
\renewcommand{\arraystretch}{1.08}
\begin{tabular*}{\textwidth}{@{\extracolsep{\fill}}lrrrrr@{}}
\toprule
Outcome & Wald OR & Wald 95\% CI & Boot OR & Boot 95\% CI & Width ratio \\
\midrule
\multicolumn{6}{@{}l}{\textit{Behavioral tier}}\\
\texttt{conflict\_frame} & 0.614 & [0.568, 0.663] & 0.615 & [0.572, 0.652] & 0.85$\times$ \\
\texttt{aggressor} & 0.732 & [0.668, 0.803] & 0.733 & [0.636, 0.814] & 1.32$\times$ \\
\texttt{strategic\_game\_frame} & 1.414 & [1.345, 1.487] & 1.411 & [1.226, 1.635] & 2.88$\times$ \\
\midrule
\multicolumn{6}{@{}l}{\textit{Moral-accounting tier and additional constructs}}\\
\texttt{leader} & 0.907 & [0.863, 0.952] & 0.907 & [0.821, 0.986] & 1.85$\times$ \\
\texttt{normaliz} & 1.080 & [1.029, 1.134] & 1.080 & [0.799, 1.249] & 4.29$\times$ \\
\texttt{target} & 1.048 & [0.995, 1.105] & 1.048 & [0.875, 1.354] & 4.36$\times$ \\
\texttt{blame\_frame} & 0.966 & [0.887, 1.052] & 0.965 & [0.748, 1.541] & 4.81$\times$ \\
\texttt{delegit} & 1.291 & [1.223, 1.363] & 1.297 & [0.999, 2.003] & 7.17$\times$ \\
\bottomrule
\end{tabular*}
\caption{Cluster-bootstrapped confidence intervals (2,000 iterations, resampling 9 outlet families with replacement) versus Wald intervals from standard MLE logit. The behavioral tier (conflict, strategic-game, aggressor) retains significance under outlet clustering with minimal to moderate CI widening; the moral-accounting tier (delegit, blame, target) loses significance under the bootstrap, reflecting outlet-structured variance that standard Wald intervals treat as independent. Width ratio = bootstrapped CI width / Wald CI width.\label{tab:bootstrap-ci}}
\end{table*}

\section{Permutation Test}
\label{app:permutation}

We shuffled LFI/RN party labels 10,000 times (holding frames, outlets, years fixed) and recomputed log-odds from the raw 2$\times$2 contingency table (Haldane correction) for each permutation. The observed conflict OR (0.594) and strategic-game OR (1.431) fall at the 0th and 100th percentiles of the null; the 99th-percentile null reaches only 1.094 and 1.058, respectively. Zero of 10,000 permutations reproduced either observed effect (two-tailed $p < 0.0001$, resolution floor). The permutation statistic is covariate-free; main-text logit uses fixed effects; substantive conclusions are identical.

\section{Multiple Comparisons Correction}
\label{app:multiple-corrections}

Table~\ref{tab:multiple-corrections} applies Bonferroni and BH ($q = 0.05$) corrections to the eight pooled logistic tests in Table~\ref{tab:main-results}. Six of eight survive both. The two non-survivors (\texttt{blame\_frame}, $p = 0.426$; \texttt{target}, $p = 0.077$) were already non-significant.

\begin{table*}[t]
\centering
\footnotesize
\setlength{\tabcolsep}{4pt}
\renewcommand{\arraystretch}{1.08}
\begin{tabular*}{\textwidth}{@{\extracolsep{\fill}}lrrllll@{}}
\toprule
Outcome & OR & Raw \emph{p} & Bonf.\ adj.\ \emph{p} & Bonf.\ sig. & BH adj.\ \emph{p} & BH sig. \\
\midrule
\multicolumn{7}{@{}l}{\textit{Survive both corrections}}\\
\texttt{conflict\_frame}        & 0.614 & $<$0.001 & $<$0.001 & *** & $<$0.001 & *** \\
\texttt{strategic\_game\_frame} & 1.414 & $<$0.001 & $<$0.001 & *** & $<$0.001 & *** \\
\texttt{delegit}                & 1.291 & $<$0.001 & $<$0.001 & *** & $<$0.001 & *** \\
\texttt{aggressor}              & 0.732 & $<$0.001 & $<$0.001 & *** & $<$0.001 & *** \\
\texttt{leader}                 & 0.907 & $<$0.001 & $<$0.001 & *** & $<$0.001 & *** \\
\texttt{normaliz}               & 1.080 & 0.002 & 0.016 & * & 0.003 & ** \\
\midrule
\multicolumn{7}{@{}l}{\textit{Do not survive (already ns in Table~\ref{tab:main-results})}}\\
\texttt{blame\_frame}           & 0.966 & 0.426 & 1.000 & ns & 0.426 & ns \\
\texttt{target}                 & 1.048 & 0.077 & 0.615 & ns & 0.088 & ns \\
\bottomrule
\end{tabular*}
\caption{Multiple comparisons correction for the eight pooled logistic regression tests in Table~\ref{tab:main-results}. Bonferroni multiplies each raw $p$ by 8 (capped at 1.0). BH applies Benjamini-Hochberg at $q = 0.05$. *** $p < 0.001$; ** $p < 0.01$; * $p < 0.05$; ns = not significant.\label{tab:multiple-corrections}}
\end{table*}

\section{BOTH-Headline Sensitivity Analysis}
\label{app:both-sensitivity}

The main analysis excludes the 1,081 \textit{BOTH} headlines (LFI and RN mentioned simultaneously or in opposition). This appendix verifies that exclusion does not drive the findings.

\paragraph{Descriptive frame rates.} BOTH headlines show elevated conflict (24.2\% vs.\ LFI 13.8\%/RN 8.7\%) and strategic-game framing (59.3\% vs.\ 37.2\%/45.9\%), and lower leader framing (25.8\% vs.\ 51.4\%/47.7\%). Kruskal--Wallis tests confirm significant differences on all outcomes except blame ($p > .05$; Table~\ref{tab:both-rates}).

\begin{table*}[t]
\centering
\footnotesize
\setlength{\tabcolsep}{4pt}
\renewcommand{\arraystretch}{1.08}
\begin{tabular*}{\textwidth}{@{\extracolsep{\fill}}lrrrr@{}}
\toprule
Outcome & LFI (\emph{n}=12,741) & RN (\emph{n}=14,770) & BOTH (\emph{n}=1,081) & KW \\
\midrule
\texttt{conflict\_frame} & 0.138 & 0.087 & 0.242 & *** \\
\texttt{strategic\_game\_frame} & 0.372 & 0.459 & 0.593 & *** \\
\texttt{delegit} & 0.257 & 0.312 & 0.327 & *** \\
\texttt{normaliz} & 0.437 & 0.451 & 0.398 & *** \\
\texttt{leader} & 0.514 & 0.477 & 0.258 & *** \\
\texttt{target} & 0.301 & 0.315 & 0.232 & *** \\
\texttt{aggressor} & 0.086 & 0.063 & 0.089 & *** \\
\texttt{blame\_frame} & 0.087 & 0.087 & 0.081 & ns \\
\bottomrule
\end{tabular*}
\caption{Frame prevalence by party group. KW = Kruskal--Wallis omnibus test across all three groups.\label{tab:both-rates}}
\end{table*}

\paragraph{Three-way regression sensitivity.} We re-estimated with a three-level specification (LFI reference; \texttt{party\_rn}$=1$; \texttt{party\_both}$=1$) using the same fixed effects. Table~\ref{tab:both-stability} shows the \texttt{party\_rn} ORs from both models; across all eight outcomes, $|\Delta| \leq 0.004$ with no direction reversals.

\begin{table*}[t]
\centering
\footnotesize
\setlength{\tabcolsep}{4pt}
\renewcommand{\arraystretch}{1.08}
\begin{tabular*}{\textwidth}{@{\extracolsep{\fill}}lrrrrc@{}}
\toprule
Outcome & 2-way OR & 3-way OR & 3-way 95\% CI & $\Delta$ & Verdict \\
\midrule
\texttt{conflict\_frame}        & 0.614 & 0.616 & [0.571, 0.666] & $+$0.002 & Robust \\
\texttt{strategic\_game\_frame} & 1.414 & 1.413 & [1.344, 1.485] & $-$0.001 & Robust \\
\texttt{delegit}                & 1.291 & 1.295 & [1.227, 1.367] & $+$0.004 & Robust \\
\texttt{aggressor}              & 0.732 & 0.732 & [0.667, 0.802] & $\phantom{+}$0.000 & Robust \\
\texttt{leader}                 & 0.907 & 0.906 & [0.863, 0.952] & $-$0.001 & Robust \\
\texttt{normaliz}               & 1.080 & 1.078 & [1.027, 1.132] & $-$0.002 & Robust \\
\texttt{target}                 & 1.048 & 1.050 & [0.996, 1.106] & $+$0.002 & Robust \\
\texttt{blame\_frame}           & 0.966 & 0.969 & [0.890, 1.055] & $+$0.003 & Robust \\
\bottomrule
\end{tabular*}
\caption{Stability of the \texttt{party\_rn} coefficient when BOTH headlines are included as a third category. 2-way OR = main analysis (LFI vs.\ RN only); 3-way OR = coefficient from LFI-reference three-level model including BOTH rows; $\Delta$ = three-way OR $-$ two-way OR. All changes are $\leq 0.004$ in absolute value; no direction reversals occur.\label{tab:both-stability}}
\end{table*}

BOTH headlines are distinct from single-party coverage (conflict OR\,=\,2.08 [1.80, 2.42]; strategic-game OR\,=\,2.39 [2.10, 2.72] vs.\ LFI) but do not confound the LFI/RN asymmetry.

\section{Annotation Error Analysis: Schema Boundary Cases}
\label{appendix-error-analysis}

The four cases below are drawn from the human audit sample (N = 400): headlines where both human coders agreed with each other and with the adjudicated label, but disagreed with the majority-vote label. Each illustrates a distinct failure mode across four different schema fields.

\begin{table*}[t]
\centering
\scriptsize
\setlength{\tabcolsep}{4pt}
\renewcommand{\arraystretch}{1.25}
\begin{tabularx}{\textwidth}{@{}>{\RaggedRight\arraybackslash}p{5.0cm}>{\RaggedRight\arraybackslash}p{1.0cm}>{\RaggedRight\arraybackslash}p{1.9cm}>{\RaggedRight\arraybackslash}p{1.0cm}>{\RaggedRight\arraybackslash}p{1.0cm}>{\RaggedRight\arraybackslash}X@{}}
\toprule
Headline & Party & Field & Human & MV & Failure mode \\
\midrule
\textit{Mélenchon voit dans le meeting du RN un «rassemblement de fachos»}\newline{\textit{[Mélenchon calls the RN rally ``a gathering of fascists'']}} &
BOTH &
\texttt{conflict} &
TRUE &
FALSE &
Mélenchon explicitly characterises the RN gathering as fascist. The model fails to transfer the aggression from inside the attributed-speech clause (``voit dans\ldots{} un'') to the conflict-frame signal for the party. \\
\addlinespace
\textit{Mosquées taguées dans le Doubs : un ex-candidat RN avoue}\newline{\textit{[Mosques tagged in the Doubs: a former RN candidate confesses]}} &
RN &
\texttt{blame} &
TRUE &
FALSE &
An RN-affiliated person confesses to vandalising mosques. Both coders assign blame; the model assigns none, likely because ``ex-candidat'' weakens the party-affiliation link below the model's attribution threshold. \\
\addlinespace
\textit{Dédiabolisation ? Marine Le Pen face au choix de la radicalité}\newline{\textit{[De-demonization? Marine Le Pen facing the choice of radicalism]}} &
RN &
\texttt{strategic\_game} &
TRUE &
FALSE &
\textit{Dédiabolisation} is the canonical French term for Le Pen's party-normalisation strategy; the headline is entirely about strategic positioning. The model misclassifies ``radicalité'' as an ideological descriptor rather than a strategic option. \\
\addlinespace
\textit{Où est passée Marine Le Pen ?}\newline{\textit{[Where has Marine Le Pen gone?]}} &
RN &
\texttt{actor\_role} &
TARGET &
NEUTRAL &
A direct question foregrounding Le Pen as the object of journalistic scrutiny. Both coders assign TARGET; the model reads the interrogative form as neutral factual inquiry rather than a targeting frame. \\
\bottomrule
\end{tabularx}
\caption{Four boundary cases from the human audit where both coders and the adjudicated label agreed but disagreed with the majority-vote label.\label{tab:error-analysis}}
\end{table*}

\section{Sentiment vs.\ Role Annotation: Three-Model Comparison}
\label{app:sentiment-robustness}

\begin{table*}[t]
\centering\small
\begin{tabular*}{\textwidth}{@{\extracolsep{\fill}}lcccc@{}}
\toprule
Model & Neg.\ base rate & Sentiment OR & 95\% CI & Conflict OR \\
\midrule
Mistral Large 2   & 63.1\% & 0.716*** & [0.677, 0.756] & 0.533*** \\
GPT-OSS-120B      & 50.3\% & 0.925**  & [0.878, 0.974] & 0.688*** \\
Llama-3.3-70B     & 35.8\% & 0.889*** & [0.842, 0.938] & 0.626*** \\
\bottomrule
\end{tabular*}
\caption{Post-hoc sentiment annotation on all party-targeted headlines with BOTH excluded ($n = 27{,}511$), compared with conflict-frame ORs from the main analysis. Sentiment base rates and ORs vary substantially across models; conflict-frame ORs are stable across all three.\label{tab:sentiment-robustness}}
\end{table*}

\end{document}